\documentclass{article}

\usepackage{arxiv}
\usepackage{xspace}
\usepackage{soul}
\usepackage{amsmath,amssymb,amsfonts}
\usepackage{algorithmic}
\usepackage{graphicx}
\usepackage{textcomp}
\usepackage{graphicx}
\usepackage[table,xcdraw]{xcolor}
\usepackage{multirow}
\usepackage{ulem}
\usepackage{booktabs}
\usepackage{siunitx}
\usepackage{placeins}
\usepackage{subcaption}
\usepackage{cite}
\usepackage[nolist]{acronym}
\usepackage{xcolor}

\newcommand\extrafootertext[1]{%
    \bgroup
    \renewcommand\thefootnote{\fnsymbol{footnote}}%
    \renewcommand\thempfootnote{\fnsymbol{mpfootnote}}%
    \footnotetext[0]{#1}%
    \egroup
}

\def\BibTeX{{\rm B\kern-.05em{\sc i\kern-.025em b}\kern-.08em
    T\kern-.1667em\lower.7ex\hbox{E}\kern-.125emX}}
\begin{document}
\title{MoViAD: A Modular Library \\for Visual Anomaly Detection}

\author{
    Manuel Barusco \\ University of Padova, Italy \\ 
    \texttt{manuel.barusco@phd.unipd.it} \\ \And
    Francesco Borsatti \\ University of Padova, Italy \\ 
    \texttt{francesco.borsatti.1@phd.unipd.it} \\ \And
    Arianna Stropeni \\ University of Padova, Italy \\ 
    \texttt{arianna.stropeni@studenti.unipd.it} \\ \And
    Davide Dalle Pezze \\ University of Padova, Italy \\ 
    \texttt{davide.dallepezze@unipd.it} \\ \And
    Gian Antonio Susto \\ University of Padova, Italy \\
    \texttt{gianantonio.susto@unipd.it} \\
}

\maketitle

\def\moviad/{MoViAD}
\def\paste/{PaSTe}
\def\mobilenet/{MobileNetV2}
\def\realiad/{Real-IAD}
\def\visa/{ViSA}
\def\mvtec/{MVTecAD}

\begin{acronym}
  \acro{VAD}{Visual Anomaly Detection}
  \acro{AAD}{Audio Anomaly Detection}
  \acro{\moviad/}{Modular Visual Anomaly Detection}
  \acro{IoT}{Internet of Things}
\end{acronym}

\begin{abstract}
\ac{VAD} is a critical field in machine learning focused on identifying deviations from normal patterns in images, often challenged by the scarcity of anomalous data and the need for unsupervised training. 
To accelerate research and deployment in this domain, we introduce \ac{MoViAD}, a comprehensive and highly modular library designed to provide fast and easy access to state-of-the-art \ac{VAD} models, trainers, datasets, and VAD utilities. 
\ac{MoViAD} supports a wide array of scenarios, including continual, semi-supervised, few-shots, noisy, and many more.
In addition, it addresses practical deployment challenges through dedicated Edge and \ac{IoT} settings, offering optimized models and backbones, along with quantization and compression utilities for efficient on-device execution and distributed inference. 
\moviad/ integrates a selection of backbones, robust evaluation VAD metrics (pixel-level and image-level) and useful profiling tools for efficiency analysis.
The library is designed for fast, effortless deployment, enabling machine learning engineers to easily use it for their specific setup with custom models, datasets, and backbones. At the same time, it offers the flexibility and extensibility researchers need to develop and experiment with new methods. 
\end{abstract}


\section{Introduction}

Visual Anomaly Detection (VAD) has emerged as a prominent area of research within the machine learning community, aiming to differentiate between normal and abnormal images within a dataset and to pinpoint the specific pixels within the image responsible for the anomaly. 
Traditional supervised learning methods often fall short in this context due to the scarcity and variability of representative samples from the anomalous class. To overcome this limitation, many techniques are designed to be trained exclusively on normal data, learning its underlying distribution and identifying deviations from that as potential anomalies.
The nature of the VAD problem makes it extremely useful and relevant in many real-world domains, such as automatic industrial inspection, healthcare diagnostics, robotics, and many others \cite{mvtec} \cite{bmad} \cite{chan2021segmentmeifyoucan}.    

In the last years a lot of research has focused on VAD, specifically for the challenging unsupervised scenario, obtaining excellent results.
However, in real-world scenarios, VAD could be evaluated in a different setting that differs from the one studied in literature.

For example, some works started to treat the VAD problem beyond the classic unsupervised setting, considering cases where a single model needs to address multiple classes (multi-class) \cite{roads} or the model needs to adapt to new items (continual) \cite{bugarin2024unveiling} \cite{oner} \cite{barusco2025memoryefficientcontinuallearning}.
Similarly, some recent works started to address the need of deployment on resource-constrained devices (edge or IoT scenario) \cite{Barusco_2025_CVPR} \cite{stropeni2025scalableiotdeploymentvisual}, while others considered the possibility of having zero or few samples of the normal class (zero-shot and few-shot settings) \cite{fewshot} \cite{fewshot1}.

Therefore, to make research easier, provide fast and easy access to these models, and navigate better among all these VAD settings, we develop the MoViAD (Modular Visual Anomaly Detection) library. 
MoViAD is a totally modular library that gives access to models, trainers, datasets, and other VAD utilities in a very modular way, allowing researchers or machine learning engineers to access what they need in a comfortable and fast way without searching for papers, code or extracting the code from other libraries. 
Moreover, practitioners can use MoViAD to quickly and effortlessly develop their specific solution, choosing between multiple components to fit their specific setting.

The rest of the paper is organized as follows: in Section \ref{sec:principles} the design principles of MoViAD are reported, and in Section \ref{sec:modules} the MoViAD library modules are listed and explained.

\section{Design principles}
\label{sec:principles}
\subsection{Modularity}
The most important design principle in MoViAD is modularity. The whole library is organised with atomic and plug-and-play modules that, when connected together, allow training and testing of the VAD models. This structure allows the researchers to use and modify the modules that are needed for their purposes and allows them to shape the training and testing flows in a custom way. 
This modular structure of MoViAD significantly benefits debugging and testing processes. The available modules in the library implement different datasets, models, trainers and other VAD utilities as reported in Section \ref{sec:modules}.

\subsection{Extensibility and Flexibility}
The modular structure of MoViAD allows the whole library to be extensible and very flexible. The library can be extended and modified in the execution flow by adding the needed modules with minimal effort and without respecting too many development constraints or internal code structures. Furthermore, connectors and adapting procedures from other libraries are available for using models available in it in \moviad/. 

\subsection{Reproducibility}
One of the objective of the MoViAD library is to facilitate comparative evaluations of various cutting-edge anomaly detection algorithms using both publicly available and custom-designed benchmark datasets. To achieve accurate and meaningful comparisons, models implementations within the MoViAD library are designed to replicate the results reported in the original research papers.

\section{Modules}
\label{sec:modules}

MoViAD library is composed of several core modules, each responsible for a distinct stage of the VAD pipeline: from dataset handling to model training, compression strategies, evaluation, and other utilities. Below, we describe the principal modules. Every module is designed to be flexible and extensible while integrating seamlessly with the other modules provided by MoViAD. 
Notably, we try to keep modules as standalone as possible so that they can be used individually inside other projects.

\subsection{Datasets}

MoViAD supports a diverse range of openly available VAD datasets. 
These datasets span several domains, from industrial to medical, including both structural and logical anomalies.
Each dataset is integrated with a unified preprocessing pipeline, standard splits, and anomaly labeling to ensure reproducibility. 
The datasets supported by MoViAD at the time of writing include:
\begin{itemize}
    \item MVTec-AD \cite{mvtec}: A widely adopted benchmark dataset for industrial anomaly detection, with pixel-level ground truth for 15 object and texture categories.
    \item ViSA \cite{visa}: Another widely adopted benchmark dataset for industrial anomaly detection, with pixel-level and image level ground truth.
    \item Real-IAD \cite{realiad}: A new real-world industrial dataset for anomaly detection.
    \item MIIC \cite{miic}: Dataset of real microscopic images of integrated circuits (ICs).
    \item MVTec-LOCO \cite{loco}: A widely adopted benchmark dataset for logical anomaly detection.
    \item BMAD \cite{bmad}: Standard dataset for VAD in the medical domain.
\end{itemize}

\subsection{VAD Methods}
This module includes a growing collection of state-of-the-art and baseline anomaly detection methods.
The currently available methods, divided by category, are: 
\begin{itemize}
    \item Memory bank-based methods: PatchCore \cite{patch} , CFA \cite{lee2022cfa} and PaDiM \cite{PaDiM}.
    \item Student-Teacher based methods: STFPM \cite{st_pyramid}, PaSTe \cite{Barusco_2025_CVPR} and RD4AD \cite{rd4ad}.
    \item Reconstruction-based methods: DRÆM~\cite{draem} and Ganomaly \cite{akcay2019ganomaly}.
    \item Normalizing flow-based methods: FastFlow \cite{yu2021fastflow}.
    \item Adversarial methods: SuperSimpleNet \cite{rolih2025supersimplenet}.
\end{itemize}
Each method integrates seamlessly with trainers and evaluation routines provided by MoViAD.

\subsection{Trainers}
The trainers module manages the full training cycle for anomaly detection methods. 
Each method is paired with a specific trainer that implements the training procedure. 
Any training procedure allows logging, checkpointing, and early stopping.

\subsection{Backbones}
This module provides access to a diverse set of backbone architectures for feature extraction.
Available backbones include: ResNet18, WideResNet, MobileNetV1 and V2, PhiNet, MCUNet and MicroNet. 
All backbones expose hooks, which make intermediate layer outputs easily configurable to enable multi-scale feature extraction. 
Also, the backbones can be trimmed to the last feature extraction layer for efficiency.
There is full compatibility with PyTorch HUB and Hugging Face Hub available models.
Every backbone that is not available in cloud download services can be implemented by adding its code and weights to the library.

\subsection{Quantization and Compression Module}
The quantization and compression module addresses storage, deployment, and communication efficiency challenges in VAD methods. This module allows:

\begin{itemize}
    \item Model quantization: Quantization of neural network weights for edge deployment. This module allows for quantization-aware training and post-training quantization of the VAD models.
    \item Feature quantization and compression: Applied to intermediate feature maps to reduce memory usage or transmission bandwidth (e.g., in federated and IoT settings).
\end{itemize}
These tools are compatible with methods and training routines and allow trade-offs between performance and resource constraints.

\subsection{Evaluation Module}
The evaluation module supports rigorous and consistent assessment of VAD models using both pixel-level and image-level metrics.
The available metrics, divided by category, are:

\begin{itemize}
    \item Pixel-level: ROC-AUC, PR-AUC, F1-score and PRO metrics for segmentation quality. 
    \item Image-level: ROC-AUC, PR-AUC and F1-score.
\end{itemize}

\subsection{Other VAD Utilities}
To streamline VAD experimentation, \moviad/ provides a set of utilities and tools:
\begin{itemize}
    \item Anomaly map postprocessing: Smoothing, normalization and thresholding for producing the final anomaly map.
    \item Data augmentation: For synthetic anomaly generation (CutPaste~\cite{li2021cutpaste}, noise injection, spatial deformation).
    \item VLM access utility from OLLAMA.
    \item Logging with Wandb.
    \item Profiling utilities for tracking model size, memory usage, and FLOPS during training and inference.
\end{itemize}

\section{Scenarios}
\moviad/ is designed to offer researchers and engineers a versatile framework for exploring various \ac{VAD} settings. Beyond providing common \ac{VAD} models and backbones, we emphasize the importance of diverse testing environments to evaluate model performance and robustness. The following paragraphs detail the key scenarios supported by \moviad/.

\subsection{Unsupervised, Noisy}
This scenario represents the standard industrial anomaly detection case, acknowledging that real-world training data may exhibit varying degrees of contamination. The core of this scenario involves training a model primarily on normal images, enabling it to learn the underlying distribution of anomaly-free samples and subsequently identify deviations during inference. 
We generalize the classic unsupervised setting by introducing a \textit{Noisy} option, allowing users to control the level of anomalous sample contamination within the training data. Contamination can be regulated at either the \textit{image-level} or \textit{pixel-level}.

Image-level contamination is implemented by adding $M$ anomalous images to $N$ normal images in the training set, such that $\frac{M}{N+M} \approx C$, where $C$ is the desired contamination proportion. Pixel-level contamination, on the other hand, requires the ground truth anomaly mask for each anomalous image. In this case, contamination is defined as $\frac{M_{pxl}}{N_{pxl} + M_{pxl}}$, where $M_{pxl}$ is the total number of anomalous pixels across all added anomalous images, and $N_{pxl}$ is the total number of pixels in the normal images. This scenario is particularly oriented towards real-world and industrial applications, where training sets may not be perfectly curated, and it is especially valuable for testing the robustness of algorithms to outliers within the training data.

\subsection{Continual Learning}
The \textit{Continual} scenario extends \moviad/'s capabilities to support training models within a continual learning paradigm. In this setting, not all tasks are available in the initial training set; instead, the model must efficiently update and train on new tasks while preserving performance on previously learned tasks. \moviad/ will implement multiple continual learning methods designed for VAD.
For example, it will consider the Replay approach.
Replay is often the most applicable out-of-the-box solution for many \ac{VAD} methods and models, since it primarily influences the dataset and training procedures.
Therefore, we used the same Replay approach tested in \cite{bugarin2024unveiling} and \cite{pezze2022continual}  to validate several VAD approaches such as STFPM, VAE, CAE, Ganomaly, FastFlow, and RIAD.
Moreover, we also implemented the variant of replay proposed in \cite{bugarin2024unveiling}, specifically designed to use replay for PatchCore and Padim models.

\subsection{Supervised, Semi-Supervised, Few-Shot}
Given that many common visual anomaly detection use cases involve data scarcity or high labeling costs, these scenarios are used for testing \ac{VAD} methods and models under the presence of limited amounts of labeled data. The \textit{Supervised} scenario can also be beneficial for assessing whether a method improves with access to all possible data, potentially serving as a theoretical upper bound on the method's performance. These scenarios introduce variations in data loading and training procedures.

\subsection{Edge, IoT}
The \textit{Edge} scenario dictates that the entire \ac{VAD} pipeline must operate efficiently on a resource-constrained device. This is crucial for low-latency requirements when the \ac{VAD} solution is deployed directly on industrial machinery; also for privacy concerns where data cannot be sent to the cloud; or simply to utilize a more energy-efficient model with an acceptable trade-off in downstream task performance.

This scenario primarily impacts the selection of models and backbones. 

For example, we provide several backbones that can be chosen from a range of tiny feature extractors like \mobilenet/~\cite{sandler2019mobilenetv2invertedresidualslinear}, in contrast to larger backbones such as Wide ResNet50~\cite{zagoruyko2017wideresidualnetworks}.
In addition, methods for making models "tiny" can include quantization or pruning strategies.
Generally, model weights can undergo quantization, which also affects the training procedure. Depending on whether models are pre-quantized or post-quantized, the training procedure differs. Pre-quantization usually necessitates a re-alignment step, while in post-quantization the model is directly trained with lower precision weights.

We also included the first models specifically designed for Edge, like in \cite{Barusco_2025_CVPR}, where it is proposed \paste/, an efficient version of STFPM designed for resource-constrained devices.

In the \ac{IoT} scenario \cite{stropeni2025scalableiotdeploymentvisual}, it is not assumed that the entire visual anomaly detection model or pipeline resides within the edge device. Instead, only a small part, such as the feature extractor, might be reside on the edge. 
This edge device in the \ac{IoT} scenario is typically associated with a cloud server that does not have strict computational resource limitations: this server completes the \ac{VAD} pipeline either by showing the results on a platform or communicating the inferred results back to the edge device.
Within this scenario, \moviad/ also offers methods to simulate the communication aspect between the edge node and the server, introducing metrics based on bitrate and other profiling tools.


\subsection{Supported Modalities}
\moviad/ is primarily focused on \ac{VAD}, with a strong emphasis on still images. 
However, the scope of VAD is continuously expanding, moving beyond image-based tasks to include additional vision-related domains.
Recent datasets, such as RealIAD /~\cite{realiad}, have introduced concepts like multiple viewing angles for a single object, adding complexity to the traditional single-image VAD problem. 
Furthermore, some VAD models are designed to operate on sequences of images, often extrapolated from videos \cite{videoad}, to detect anomalies in dynamic scenes. 
Currently, \moviad/ robustly supports metrics and models tailored for single still images, with future plans to extend this support to incorporate multiple viewing angles and video anomaly detection capabilities.
Another recent dataset that is pushing the boundaries of VAD is MVTec 3D-AD Dataset \cite{mvtec3d}, the first comprehensive 3D dataset for the task of unsupervised anomaly detection and localization and 3D data. In the future we are considering to incorporate 3D VAD in the library.
Beyond visual data, \moviad/ also aims to support \ac{AAD}. This initiative stems from the growing interest in adapting successful \ac{VAD} methods to tackle \ac{AAD} tasks \cite{barusco2025visionsoundadvancingaudio}. 
The typical approach involves an initial pre-processing step where the raw audio (time-series) signal is transformed into an image representation, usually through a spectrogram transform such as the Log Mel Spectrogram. 
These spectrogram images necessitate specific feature extractors that are trained on such data, distinguishing them from the general-purpose feature extractors used for natural images.

\bibliographystyle{IEEEtran}
\bibliography{references}

\end{document}